\documentclass[pdflatex,sn-mathphys-num]{sn-jnl}

\usepackage{graphicx}%
\usepackage{multirow}%
\usepackage{amsmath,amssymb,amsfonts}%
\usepackage{amsthm}%
\usepackage[title]{appendix}%
\usepackage{xcolor}%
\usepackage{textcomp}%
\usepackage{manyfoot}%
\usepackage{booktabs}%
\usepackage{algorithm}%
\usepackage{algorithmicx}%
\usepackage{algpseudocode}%
\usepackage{listings}%
\usepackage[T1]{fontenc}
\usepackage[utf8]{inputenc}
\usepackage{placeins}

\theoremstyle{thmstyleone}%
\theoremstyle{thmstyletwo}%

\theoremstyle{thmstylethree}%

\hypersetup{bookmarksdepth=2}

\begin{document}

\title[Article Title]{EEG-Based Biometric verification in Epilepsy: A Baseline Evaluation Across Multiple Stimulus Conditions}


\author[1]{\fnm{Krishna} \sur{Patel}}

\author[1]{\fnm{Narayan} \sur{Vetrekar}}

\author[1]{\fnm{Marissa L.} \spfx{de} \sur{Ataide}}

\author[2]{\fnm{Teresa} \sur{Ferreira}}

\author*[1]{\fnm{R. S.} \sur{Gad}}
\email{rsgad@unigoa.ac.in}

\author[3]{\fnm{Raghavendra} \sur{Ramachandra}}

\affil[1]{\orgdiv{School of Physical and Applied Sciences},
\orgname{Goa University},
\orgaddress{\state{Goa}, \country{India}}}

\affil[2]{\orgdiv{Neurology Department},
\orgname{Goa Medical College and Hospital},
\orgaddress{\city{Goa}, \postcode{India}}}

\affil[3]{\orgdiv{Department of Information Security and Communication Technology},
\orgname{Norwegian University of Science and Technology},
\orgaddress{\city{Gjøvik}, \country{Norway}}}

\vspace{-4mm}
\vspace{-4mm}
\abstract{Electroencephalography (EEG)-based biometrics provide a promising approach for identity verification, but the influence of epilepsy on verification performance remains insufficiently investigated. This study systematically evaluates EEG-based biometric verification in healthy participants and participants with epilepsy using a balanced database comprising 60 participants (30 healthy and 30 with epilepsy), yielding 207,360 channel-wise, non-overlapping 10-s EEG segments. Thirty-two-channel interictal EEG was recorded across diverse stimulation conditions. To examine whether epilepsy-related spectral variations influence EEG-based verification, log-absolute power features were extracted from five canonical frequency bands and evaluated using two conventional machine-learning and two deep-learning methods. A progressive evaluation framework introduced increasing mismatch between the enrolment and probe data across stimulus conditions and data-collection protocols (DCP). Performance was evaluated using EER and GMR at FMRs of 0.01\% and 0.1\%, together with condition-level statistical comparisons and participant-cluster bootstrap inference. Across both cohorts, verification performance decreased as the mismatch between enrolment and probe conditions increased, while BiLSTM demonstrated the greatest overall robustness across all three experimental settings.}

\keywords{{}EEG-based biometrics, Epilepsy, Interictal scalp EEG, BiLSTM}

\maketitle
 
\section{Introduction}\label{sec1}
\vspace{-2mm}
Biometric verification has become important for securing digital services, healthcare information, financial systems, and restricted resources. Although established traits such as the face, fingerprint, and iris provide effective identity recognition, their external visibility may permit unauthorized capture or replication. EEG offers a promising internal biometric alternative for high-security access control and identity verification in sensitive environments, because ongoing brain activity contains stable person-specific spatial, spectral, and temporal signatures suitable for identity recognition \cite{Du2022ETST,Yang2022Brainprints}.

EEG biometric information is nevertheless affected by cognitive state, external stimulation, medication, neurological condition, and recording session. A practical system must therefore preserve identity-related information when the EEG used for enrolment and verification is acquired under varying conditions. This requirement is particularly important for neurologically diverse users, whose brain activity may differ from that represented in predominantly healthy biometric databases.

Epilepsy provides an important case for such evaluation. The World Health Organization estimates that approximately 50 million people worldwide have epilepsy, with active epilepsy affecting about 4--10 per 1000 people \cite{WHO2024Epilepsy}. Epilepsy is associated with abnormal neuronal excitability and altered functional organization that may remain observable during the interictal period. Reported interictal EEG alterations include:

\begin{itemize}

\item \textit{Alpha-rhythm alterations:} Irregularity or slowing of the alpha rhythm, together with a shift from higher- to lower-frequency alpha activity \cite{Pyrzowski2015Focal};

\item \textit{Spectral-power alterations:} Increased slow-frequency activity and changes in delta, theta, beta, and gamma power, accompanied by reduced peak alpha power \cite{Pegg2020Interictal}; and

\item \textit{Functional-network alterations:} Altered synchronization, functional connectivity, and brain-network organization, including changes detectable in EEG intervals without visible epileptiform discharges \cite{Woldman2020Networks}.

\end{itemize}
These alterations overlap with the spectral and network characteristics used for EEG biometrics and may therefore influence within-person repeatability or between-person
separability. They could reduce template stability or, if sufficiently consistent within an individual, provide additional subject-specific information. Consequently, the effectiveness of established EEG biometric methods in individuals with epilepsy cannot be inferred from their performance in healthy populations or from clinical EEG alterations
alone, highlighting the need for direct evaluation across different enrolment and probe conditions.

\vspace{-2mm}
\subsection{Literature and Contributions}
\label{subsec:literature_contributions}
\vspace{-2mm}
Recent EEG biometric studies have demonstrated person-specific characteristics using neurologically healthy datasets \cite{Du2022ETST}. Resting-state EEG, particularly eyes-closed and eyes-open recordings, is widely used because it requires minimal participant effort while retaining subject-specific spectral information \cite{Lai2019RestingEEG}. Subsequent studies introduced audio-visual stimulation to evoke event-related neural responses and mental activities such as counting, motor imagery, and other cognitive tasks to enhance individual discrimination. Emotional and visual stimulation paradigms have also demonstrated promising performance for EEG-based person recognition \cite{RuizBlondet2016,Wilaiprasitporn2020}.

Different EEG representations have been investigated for biometric verification, including time-domain, frequency-domain, and time--frequency representations derived using power spectral density, wavelet transforms, and short-time Fourier transforms. These features have been evaluated using conventional classifiers such as SVM, LDA, KNN, artificial neural networks, random forests, and decision trees \cite{Shams2022Survey}. More recently, deep-learning models based on CNN and LSTM architectures have demonstrated their ability to learn
spatial and temporal identity information from EEG signals and their derived representations \cite{Wilaiprasitporn2020}.

Despite these developments, most existing systems perform enrolment and verification using EEG recorded under the same stimulus or mental task. These matched-stimulus experiments establish the repeatability of EEG biometric characteristics under controlled conditions but do not fully evaluate their robustness when the cognitive or emotional state changes. Recent cross-stimulation studies have shown that the relationship between enrolment and probe conditions can substantially affect verification performance \cite{Ataide2024}. However, these studies have also focused primarily on healthy participants, and the reliability of cross-stimulus EEG verification in users with neurological disorders remains insufficiently investigated. This gap is particularly relevant to individuals with epilepsy, whose interictal EEG characteristics may influence biometric verification
\cite{Pyrzowski2015Focal,Pegg2020Interictal,Woldman2020Networks}. Reliable EEG-based verification is therefore important in healthcare, personal-identification, and access-control applications. Accordingly, this study investigates EEG-based biometric verification in participants with epilepsy across multiple stimulation and enrolment--probe conditions. To the best of our knowledge, this is the first study to benchmark epilepsy-focused EEG biometric verification under both matched- and cross-stimulus conditions.


Owing to the absence of a suitable publicly available database, we constructed a balanced in-house EEG database comprising 60 participants (30 healthy and 30 with epilepsy), yielding 207,360 non-overlapping 10-s channel-wise EEG segments across resting-state, relaxation, emotional-video, and alphanumeric flicker conditions in two data-collection protocols (DCPs).
Motivated by epilepsy-related spectral alterations, log-absolute power features were extracted from the five canonical EEG bands \cite{Pyrzowski2015Focal,Pegg2020Interictal}. Verification was performed independently within each cohort to assess identity-related EEG repeatability and discriminability, rather than classifying neurological status.


Four algorithms---CRC, SVM, CNN, and BiLSTM---were benchmarked through three progressively challenging experiments involving intra-stimulus verification, inter-stimulus verification within the same DCP, and inter-stimulus verification across DCPs. Performance was evaluated using equal error rate (EER) and genuine match rate (GMR) at false match rates (FMRs) of 0.01\% and 0.1\%, together with participant-cluster bootstrap confidence intervals to account for repeated EEG observations within participants.

The major contributions of this study are as follows:

\begin{itemize}

\item We present a systematic performance analysis of EEG-based biometric verification for healthy participants and participants with epilepsy under identical experimental conditions. To the best of our knowledge, this is the first epilepsy-focused EEG biometric study
covering both matched- and cross-stimulus verification.

\item We introduce a newly constructed, balanced EEG-verification database comprising 60 participants and 207,360 non-overlapping 10-s EEG segments acquired under resting-state, relaxation, emotional-video, and alphanumeric flicker conditions across two DCPs.

\item We benchmark CRC, SVM, CNN, and BiLSTM independently on the healthy and epilepsy cohorts through intra-stimulus, within-DCP inter-stimulus, and cross-DCP inter-stimulus experiments. Performance is evaluated using EER and GMR, with participant-cluster bootstrap confidence intervals accounting for repeated EEG observations within participants.

\end{itemize}

The remainder of this paper is organised as follows: Section 2 describes the database and preprocessing; Section 3 presents the verification methods;
Section 4 reports the experiments and results; and Section 5 concludes the study.


\vspace{-2mm}
\section{Epilepsy Neurological (EpiNeuro) Database}
\label{sec:EpiNeuro database}
\vspace{-2mm}

The study included 60 adults aged 18--65 years, comprising 30 participants with clinically diagnosed epilepsy (15 focal and 15 generalized; recruited based on availability) and 30 matched healthy participants. Epilepsy diagnoses were documented by the treating neurologist, and participants with an additional neurological disorder were excluded. Ethical approval was obtained from the Institutional Human Ethics Committee, Goa University (GU-DRDRM/IHEC-Cert/2023/169), for healthy participants, and the Institutional Ethics Committee, Goa Medical College (GMCIEC/2024/338), for participants with epilepsy. All participants provided written informed consent after being informed of the recording protocol, including clinically reviewed stimuli submitted for ethics review. Healthy participants reported no history of epilepsy or other neurological disorders. Participants with epilepsy continued their prescribed antiseizure medication, and no clinical seizure occurred during EEG acquisition; therefore, only interictal EEG was analysed. To assess demographic comparability, the age and gender characteristics summarized in Table~\ref{tab:demographic_characteristics} were statistically examined. Because the Lilliefors test rejected age normality for the healthy group ($p=0.0136$), but not for the epilepsy group ($p=0.0516$), age was compared using a two-sided Mann--Whitney U test. No significant difference was observed in age ($U=435.5$, $p=0.836$; $r=-0.032$) or gender distribution ($\chi^{2}(1)=0.606$, $p=0.436$; Cram\'er's $V=0.101$), indicating no statistically significant demographic imbalance between the cohorts.

\begin{table}[t]
\caption{Demographic characteristics of the study participants.}
\label{tab:demographic_characteristics}
\centering
\small
\setlength{\tabcolsep}{4pt}

\begin{tabular*}{\linewidth}{@{\extracolsep{\fill}}lccccc}
\toprule
Group & $n$ & Mean $\pm$ SD & Median (IQR) & Range & Male/Female \\
\midrule
Healthy  & 30 & $33.73 \pm 12.92$ & 29.5 (19.0) & 18--65 & 12/18 \\
Epilepsy & 30 & $33.97 \pm 12.81$ & 30.0 (17.0) & 18--65 & 15/15 \\
\bottomrule
\end{tabular*}

\vspace{1mm}
{\footnotesize\raggedright
Age is reported in years; SD: standard deviation; IQR: interquartile
range.\par}

\vspace{-7mm}
\end{table}

\begin{figure}[!t]
\vspace{-2mm}
\centering
\includegraphics[width=1\textwidth]
{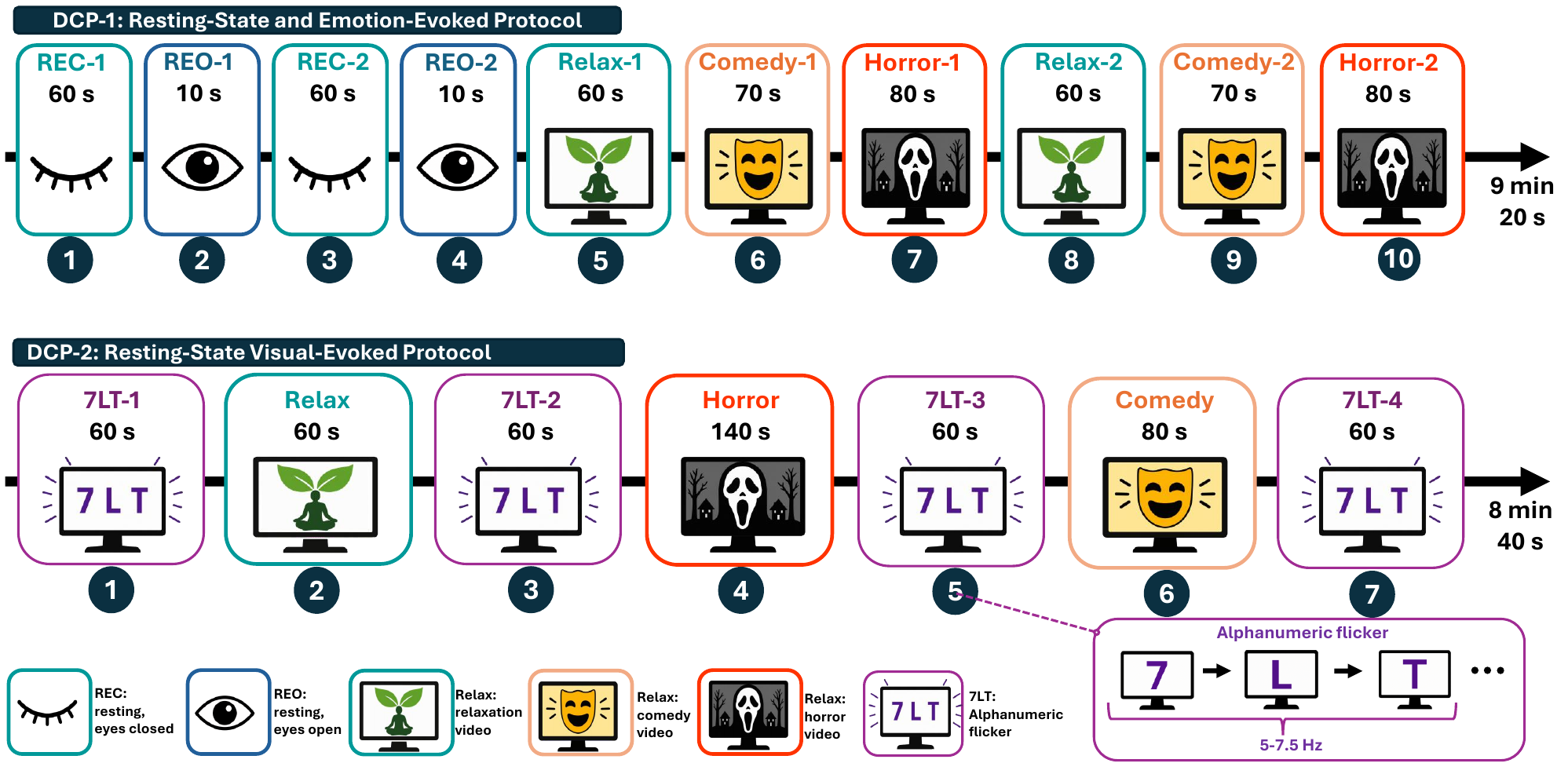}
\vspace{-2.5mm}
\caption{Timing and sequence of stimulus conditions in Data-Collection
Protocols 1 and 2.}
\label{fig:DCP}
\vspace{-2mm}
\end{figure}

\paragraph{EEG acquisition}
 EEG signals were acquired using a wireless LiveAmp~32 amplifier connected to a 32-channel R-Net electrode system (Brain Products GmbH, Germany), with BrainVision Recorder v1.22.0001 used for recording. The passive, water-based electrodes were arranged according to the international 10--20 system, and electrode impedances were maintained below $10~\mathrm{k}\Omega$. EEG signals were sampled at 250~Hz without user-defined online filtering to avoid irreversible frequency-component removal and preserve flexibility for offline preprocessing.  Identical recording equipment and settings were used for the healthy and epilepsy cohorts. Participants were seated approximately 76~cm from the stimulus display in a quiet room and instructed to remain relaxed while minimizing movement and unnecessary blinking.
\vspace{-2mm}
\paragraph{Audio Visual Stimuli}
\textit{Data-collection protocol 1:} DCP-1 combined resting-state EEG with emotion-evoked audiovisual stimulation to capture subject-specific brain activity under different levels of visual and emotional engagement. It began with a 60-s eyes-closed recording (REC1), providing a resting baseline with minimal visual input, followed by a 10-s eyes-open recording (REO1) to introduce visual engagement. These conditions were repeated as REC2 and REO2 for 60 and 10~s, respectively, enabling the assessment of within-condition EEG repeatability to assess robustness to stimulus-dependent EEG variation. Participants subsequently viewed two successive sets of relaxation, comedy, and horror videos, intended to evoke calm, positive, and negative emotional states, respectively. This variation enabled verification performance to be examined both within the same stimulus and across changing emotional conditions. The relaxation, comedy, and horror recordings lasted 60, 70, and 80~s, respectively, reflecting the lengths of the selected emotion-eliciting videos. As shown in Fig.~\ref{fig:DCP}, DCP-1 provided a total of 560~s of EEG.

\textit{Data-collection protocol 2:} DCP-2 combined alphanumeric flicker stimulation with relaxation and emotion-evoked videos to examine subject-specific EEG under sustained visual attention and changing emotional conditions. It comprised four 60-s flicker recordings, denoted 7LT1--7LT4, arranged in the sequence 7LT1, relaxation, 7LT2, horror, 7LT3, comedy, and 7LT4. The repeated flicker blocks enabled within-stimulus repeatability to be assessed at increasing temporal separations, while their interleaving with the videos supported verification across different stimulation conditions. The relaxation, horror, and comedy videos lasted 60, 140, and 80~s, respectively, reflecting the lengths of the selected emotion-eliciting videos, and provided a total of 520~s of EEG as described in Fig.~\ref{fig:DCP}. The flicker paradigm presented the characters 7, L, and T at frequencies between 5 and 7.5~Hz to elicit frequency-dependent visual responses, with an auditory cue directing attention to the target character. The paradigm was generated using MATLAB R2021a, and stimulus onset and termination boundaries were documented manually for subsequent offline segmentation.
\vspace{-2mm}
\paragraph{EEG Preprocessing and Segmentation}
Continuous EEG was processed offline using MATLAB R2025b. A Butterworth band-pass filter from 0.3 to 100~Hz and a second-order IIR notch filter centred at 50~Hz ($Q=35$) were applied forward and backward using \texttt{filtfilt} to obtain a zero-phase response. The high-pass stage attenuated the DC component and slow baseline drift, while no offline re-referencing was performed. Before spectral estimation, each channel was additionally demeaned within its corresponding analysis window, as described in Section~\ref{sec:eeg_verification}. Using the manually documented stimulus boundaries, each condition was extracted independently and divided into non-overlapping 10-s windows. Each multichannel segment contained 2500 samples from 32 channels and was represented as a $32\times2500$ matrix. DCP-1 and DCP-2 yielded 56 and 52 segments per participant, respectively, corresponding to 108 multichannel segments or 3456 channel-wise segments per participant. The two cohorts therefore contributed 103,680 [$(32\times56\times30)+ (32\times52\times30)$] channel-wise segments each, resulting in \textbf{207,360} channel-wise EEG segments overall. Identical preprocessing and segmentation procedures were applied to both cohorts.

\begin{table}[t]
\caption{Composition of the EEG database acquired under DCP-1 and DCP-2.}
\label{tab:eeg_database_composition}
\centering
\scriptsize
\setlength{\tabcolsep}{3pt}

\begin{tabular*}{\textwidth}
{@{\extracolsep{\fill}}llcccc@{}}
\toprule
\textbf{Protocol} & \textbf{Condition} & \textbf{Recordings} &
\textbf{Segments(s)/} & \textbf{Healthy} & \textbf{Epilepsy}\\
& & & \textbf{participant(n)} &
$[s \times n \times \mathrm{ch}]$ & $[s \times n \times \mathrm{ch}]$\\
\midrule

DCP-1 & Resting state & REC1, REC2
& 12 & 11,520 & 11,520\\

& Eyes open & REO1, REO2
& 2 & 1,920 & 1,920\\

& Relaxation & Relax1, Relax2
& 12 & 11,520 & 11,520\\

& Comedy & Comedy1, Comedy2
& 14 & 13,440 & 13,440\\

& Horror & Horror1, Horror2
& 16 & 15,360 & 15,360\\

\cmidrule(lr){2-6}
& \textbf{DCP-1 total} & &
\textbf{56} & \textbf{53,760} & \textbf{53,760}\\

\midrule

DCP-2 & Alphanumeric flicker & 7LT1--7LT4
& 24 & 23,040 & 23,040\\

& Relaxation & Relax
& 6 & 5,760 & 5,760\\

& Comedy & Comedy
& 8 & 7,680 & 7,680\\

& Horror & Horror
& 14 & 13,440 & 13,440\\

\cmidrule(lr){2-6}
& \textbf{DCP-2 total} & &
\textbf{52} & \textbf{49,920} & \textbf{49,920}\\

\midrule

\multicolumn{3}{l}{\textbf{Complete database}} &
\textbf{108} & \textbf{103,680} & \textbf{103,680}\\

\botrule
\end{tabular*}

\begin{flushleft}
\end{flushleft}
\vspace{-6mm}
\end{table}

\vspace{-2mm}
\section{EEG-Based User verification}
\label{sec:eeg_verification}
\begin{figure}[!t]
\vspace{-2mm}
\centering
\includegraphics[width=\textwidth]
{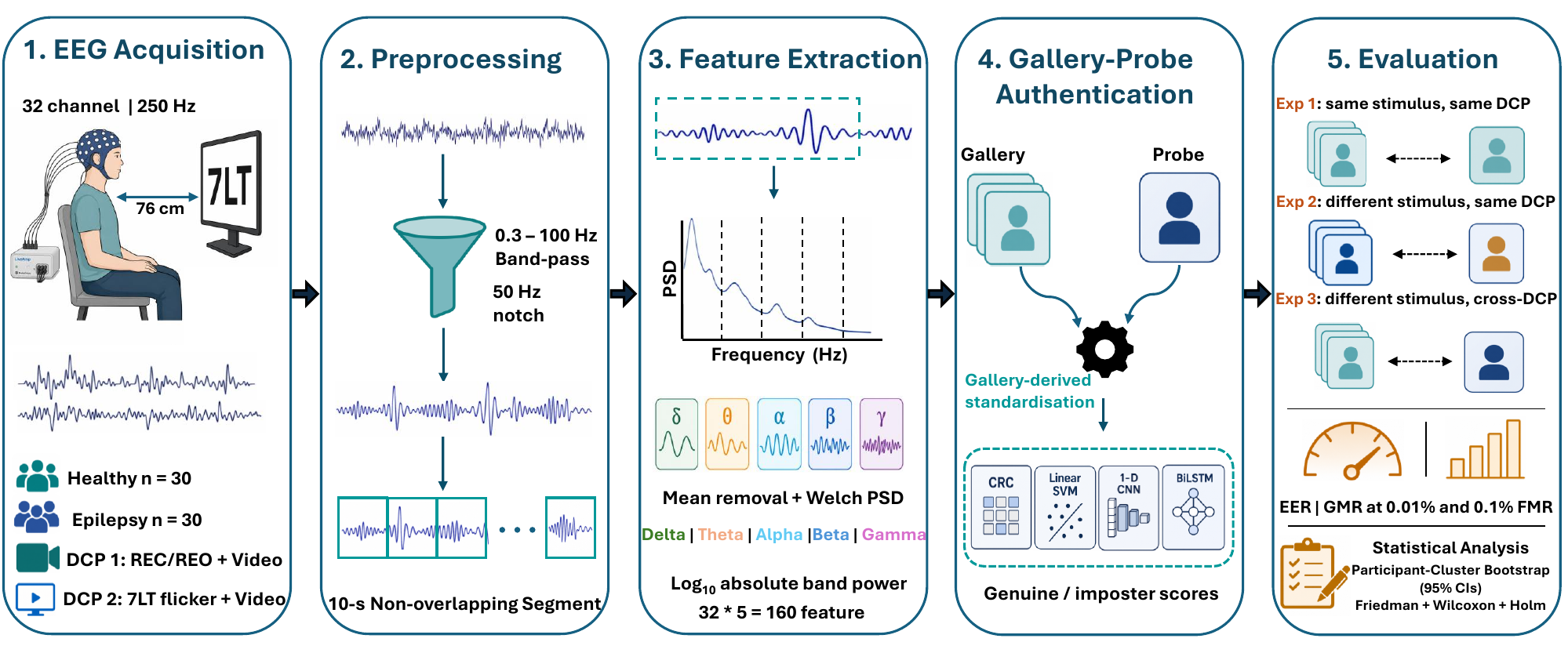}
\vspace{-1.5mm}
\caption{Overall EEG-based biometric verification framework for healthy
and epilepsy cohorts.}
\label{fig:overall_framework}
\vspace{-2mm}
\end{figure}

\vspace{-1mm}
This section presents the proposed EEG-based user verification framework using two conventional machine-learning and two deep-learning approaches. Figure~\ref{fig:overall_framework} illustrates the complete workflow, comprising feature extraction, gallery--probe construction and model training. The individual stages are described below.

\paragraph{Feature Extraction}
Let the preprocessed 10-s EEG segment obtained from participant $i$
under stimulus condition $q$ be represented as
\vspace{-2.5mm}
\begin{equation}
\mathbf{X}_{i,k}^{(q)}
=
[x_c[n]]_{C\times T}
\in\mathbb{R}^{32\times2500},
\label{eq:eeg_segment}
\end{equation}
where $k$ denotes the segment index, $C=32$ is the number of scalp EEG channels, and $T=2500$ is the number of temporal samples recorded at 250~Hz. The index $q$ identifies the stimulus condition presented in its predefined sequence within the corresponding DCP. Each
$\mathbf{X}_{i,k}^{(q)}$ was subsequently transformed into a spectral representation through \textit{feature extraction}. Log-absolute band-power features were selected because spectral representations have previously demonstrated their suitability for EEG-based verification \cite{Plucinska2023Spectral}. For a channel signal $x_c[n]$ containing $N$ samples, the temporal mean was first removed as
\vspace{-2.5mm}
\begin{equation}
\widetilde{x}_c[n]
=
x_c[n]-\frac{1}{N}\sum_{m=1}^{N}x_c[m].
\label{eq:demeaning}
\end{equation}
For SVM and CRC, the power spectral density (PSD) of each demeaned 10-s segment was estimated using Welch's method \cite{Plucinska2023Spectral}, with a 2-s Hamming window, 50\% overlap within each 10-s segment, and a 1024-point discrete Fourier transform. For channel $c$ and frequency band $b$, the log-absolute band power was calculated as
\vspace{-2.5mm}
\begin{equation}
A_{c,b}
=
\log_{10}\!\left[
\max\!\left(
\int_{f_b^{\mathrm{L}}}^{f_b^{\mathrm{U}}}
S_c(f)\,\mathrm{d}f,\epsilon
\right)\right],
\label{eq:log_band_power}
\end{equation}
where $S_c(f)$ denotes the PSD, $f_b^{\mathrm{L}}$ and $f_b^{\mathrm{U}}$ are the corresponding frequency limits, and $\epsilon$ prevents undefined logarithmic values. Five canonical bands were considered: delta ($0.5$--$4$~Hz), theta ($4$--$8$~Hz), alpha ($8$--$13$~Hz), beta ($13$--$30$~Hz), and gamma ($30$--$45$~Hz). Concatenating the five band-power values across the 32 channels produced
\begin{equation}
\mathbf{f}
=
[A_{1,\delta},\ldots,A_{32,\gamma}]^{\mathrm{T}}
\in\mathbb{R}^{160}.
\label{eq:feature_vector}
\end{equation}
This whole-segment vector provided a compact fixed-dimensional representation for SVM and CRC. For CNN and BiLSTM, the temporal order within each segment was retained so that the models could learn short- and long-term variations in spectral activity \cite{Du2022ETST}. Accordingly, each 10-s segment was divided into ten consecutive, non-overlapping 1-s windows. Every window was independently demeaned and analysed using a 250-sample Hamming window and a 512-point discrete Fourier transform. The resulting 160-dimensional vectors were arranged in temporal order as 
\vspace{-2.5mm}
\begin{equation}
\mathbf{F}
=
[\mathbf{f}_1,\mathbf{f}_2,\ldots,\mathbf{f}_{10}]
\in\mathbb{R}^{160\times10}.
\label{eq:feature_sequence}
\end{equation}
Consequently, SVM and CRC received a $160\times1$ whole-segment vector, whereas CNN and BiLSTM received a $160\times10$ spectral sequence. Despite this difference in input organisation, all models used the same 10-s EEG segments, five-band spectral information, gallery--probe partitions, and evaluation criteria. Thus, the results compare complete architecture-specific verification pipelines under consistent experimental conditions.

Following feature extraction, the \textit{gallery--probe Experimental protocol} organized the resulting representations into recording-defined enrolment and testing sets without random mixing of segments. For example, in the REC1--REC2 comparison, REC1 was used exclusively as the
gallery and REC2 exclusively as the probe. For feature dimension $d$, gallery-derived standardization was performed as:

\vspace{-2.5mm}
\begin{equation}
\widehat{x}_{i,d}
=
\frac{x_{i,d}-\mu_d^{\mathcal{G}}}
     {\sigma_d^{\mathcal{G}}},
\label{eq:gallery_standardization}
\end{equation}
\vspace{-2.5mm}

where $\mu_d^{\mathcal{G}}$ and $\sigma_d^{\mathcal{G}}$ were estimated exclusively from the gallery and subsequently applied to both gallery and probe observations. Gallery standard deviations below $10^{-10}$ were replaced by one to prevent division by zero. This procedure prevented data leakage by ensuring that probe information did not influence feature standardization or model training.
\vspace{-2mm}
\paragraph{Verification models}
The standardized representations were evaluated using four \textit{verification models}: CRC, linear SVM, 1-D CNN, and BiLSTM. Identical model configurations were maintained across the stimulus comparisons and cohorts. A separate model was trained for each gallery--probe comparison using only its corresponding gallery samples.

\textit{Linear ECOC-SVM:} Each standardized 160-dimensional feature vector was classified using linear SVM learners arranged in a one-versus-one error-correcting output-code framework \cite{Ataide2024}. For 30 enrolled identities, this resulted in
435 binary classifiers. The class scores generated for each probe were used for rank-1 identity prediction and verification-score formation.

\textit{Collaborative representation classifier:} CRC employed regularized least-squares collaborative representation \cite{Zhang2011CRC}. A gallery-derived eigenfeature projection reduced the standardized input from 160 to 24 dimensions, after which gallery and probe vectors were normalized to unit $\ell_2$ norm. Collaborative coefficients were estimated using an $\ell_2$ regularization parameter of $\kappa=0.001$, and each probe was assigned to the identity producing the minimum class-specific reconstruction distance. For verification, the distances were transformed as
$s_c=-\log[\max(r_c,\epsilon)]$, ensuring that higher scores represented
stronger matches.

\textit{One-dimensional CNN:} The CNN received the $160\times10$ temporally ordered spectral sequence. Its architecture comprised two causal one-dimensional convolutional layers containing 16 and eight filters of length five, respectively. Each convolution was followed by
rectified linear-unit activation and layer normalisation \cite{Lai2022CNN}. Global average pooling summarised the temporal activations, while a 30-unit fully connected layer followed by softmax classification generated the identity scores. The CNN was trained exclusively on gallery samples for 50 epochs using stochastic gradient descent with momentum, an initial learning rate of 0.01, a mini-batch size of 35, and a gradient threshold of one. Probe samples were reserved exclusively for verification evaluation.

\textit{Bidirectional LSTM:} The BiLSTM received the same $160\times10$ spectral sequence and processed its ten temporal windows in both the forward and backwards directions, consistent with recent BiLSTM-based EEG biometric research \cite{Mishra2024CNNBiLSTM}. The network contained 70 hidden units in each direction and used its final 140-dimensional bidirectional representation for classification. A 30-unit fully connected layer followed by softmax classification generated the identity scores. The BiLSTM was trained exclusively on gallery samples for 50 epochs using stochastic gradient descent with momentum, an initial learning rate of 0.01, a mini-batch size of 35, and a gradient threshold of one. Probe samples were used only for verification evaluation.
\vspace{-2mm}

\section{Results and Discussion}
\label{sec:results_discussion}
\vspace{-2mm}
This section presents the experimental evaluation and corresponding results for EEG-based user verification in the healthy and epilepsy cohorts. CRC, SVM, CNN, and BiLSTM were evaluated using equal error rate (EER) and genuine match rate (GMR) at false match rates (FMRs) of $0.01\%$ and $0.1\%$, where lower EER and higher GMR indicate better verification performance.

To examine robustness under increasing variation, three experiments progressively increased the mismatch between the enrolment and probe EEG. \textbf{Experiment~1} used separate recordings of the same stimulus within a DCP to establish baseline repeatability. \textbf{Experiment~2} compared
different stimuli within the same DCP to assess within-DCP stability, whereas \textbf{Experiment~3} compared different stimuli across DCP-1 and DCP-2 to assess cross-DCP generalization. All experiments were conducted independently for the two cohorts using identical features, gallery--probe separation, model configurations, balanced segment counts for corresponding cohort comparisons, and evaluation procedures, thereby enabling a consistent comparison of method and cohort effects.

\begin{table*}[t]
\caption{Experiment 2: inter-stimulus verification within each
data-collection protocol. All values are in \%; GMR subscripts denote
the corresponding FMR operating points.}
\label{tab:experiment2_results}%
\centering
\scriptsize
\setlength{\tabcolsep}{1.5pt}
\renewcommand{\arraystretch}{1.02}

\begin{tabular*}{\textwidth}
{@{\extracolsep\fill}llll*{8}{c}@{}}
\toprule
\multirow{2}{*}{DCP} &
\multirow{2}{*}{Enrolment} &
\multirow{2}{*}{Probe} &
\multirow{2}{*}{Metric} &
\multicolumn{4}{c}{Healthy subjects} &
\multicolumn{4}{c}{Seizure subjects} \\
\cmidrule(lr){5-8}\cmidrule(lr){9-12}
& & & &
SVM & CRC & LSTM & CNN &
SVM & CRC & LSTM & CNN \\
\midrule

\multirow{18}{*}{DCP-1}
& \multirow{9}{*}{REC 1,2}
& \multirow{3}{*}{Relax 1,2}
& EER
& 9.91 & 5.57 & 2.21 & 7.46
& 6.08 & 3.58 & 0.55 & 3.62 \\

& & & GMR$_{0.01}$
& 6.39 & 60.56 & 73.33 & 38.89
& 30.83 & 76.39 & 94.44 & 57.78 \\

& & & GMR$_{0.1}$
& 30.83 & 70.28 & 80.83 & 56.11
& 46.94 & 83.06 & 98.33 & 74.44 \\
\cmidrule(lr){3-12}

&
& \multirow{3}{*}{Comedy 1,2}
& EER
& 12.64 & 6.97 & 2.18 & 5.93
& 6.43 & 4.68 & 1.90 & 6.53 \\

& & & GMR$_{0.01}$
& 11.67 & 35.95 & 59.76 & 47.38
& 19.52 & 25.48 & 68.33 & 37.14 \\

& & & GMR$_{0.1}$
& 25.71 & 57.14 & 71.43 & 64.52
& 38.81 & 69.52 & 84.76 & 65.48 \\
\cmidrule(lr){3-12}

&
& \multirow{3}{*}{Horror 1,2}
& EER
& 11.67 & 7.15 & 2.49 & 8.46
& 6.20 & 6.05 & 1.65 & 5.64 \\

& & & GMR$_{0.01}$
& 7.50 & 47.71 & 59.79 & 26.67
& 17.50 & 28.54 & 69.17 & 44.58 \\

& & & GMR$_{0.1}$
& 25.63 & 59.58 & 69.79 & 52.71
& 29.17 & 46.46 & 84.38 & 58.75 \\
\cmidrule(lr){2-12}

& \multirow{9}{*}{Relax 1,2}
& \multirow{3}{*}{REC 1,2}
& EER
& 9.90 & 7.69 & 1.12 & 2.71
& 6.09 & 3.84 & 1.94 & 3.91 \\

& & & GMR$_{0.01}$
& 7.22 & 60.83 & 93.06 & 76.67
& 12.78 & 54.72 & 62.50 & 27.22 \\

& & & GMR$_{0.1}$
& 27.50 & 71.11 & 96.11 & 88.33
& 50.28 & 69.17 & 90.00 & 55.28 \\
\cmidrule(lr){3-12}

&
& \multirow{3}{*}{Comedy 1,2}
& EER
& 1.90 & 1.92 & 0.24 & 0.71
& 0.96 & 0.53 & \textcolor{blue}{0.00} & 1.42 \\

& & & GMR$_{0.01}$
& 61.67 & 70.71 & 97.86 & 94.76
& 79.52 & 90.24 & \textcolor{blue}{100.00} & 73.33 \\

& & & GMR$_{0.1}$
& 86.90 & 82.38 & 99.76 & 98.10
& 88.33 & 96.19 & \textcolor{blue}{100.00} & 97.62 \\
\cmidrule(lr){3-12}

&
& \multirow{3}{*}{Horror 1,2}
& EER
& 2.04 & 1.24 & \textcolor{blue}{0.01} & 0.24
& 1.25 & 2.08 & 0.21 & 1.94 \\

& & & GMR$_{0.01}$
& 69.79 & 73.33 & \textcolor{blue}{99.58} & 96.67
& 66.04 & 89.79 & 98.75 & 91.04 \\

& & & GMR$_{0.1}$
& 91.04 & 89.38 & \textcolor{blue}{100.00} & 99.38
& 87.50 & 93.33 & 99.79 & 95.83 \\
\midrule

\multirow{18}{*}{DCP-2}
& \multirow{9}{*}{L7T 1}
& \multirow{3}{*}{Relax}
& EER
& 3.32 & 2.76 & 0.50 & 1.53
& 4.31 & 1.67 & \textcolor{blue}{0.00} & 2.23 \\

& & & GMR$_{0.01}$
& 2.22 & 73.89 & 86.11 & 67.22
& 69.44 & 81.67 & \textcolor{blue}{100.00} & 81.11 \\

& & & GMR$_{0.1}$
& 66.67 & 93.33 & 98.89 & 93.33
& 82.22 & 93.89 & \textcolor{blue}{100.00} & 91.67 \\
\cmidrule(lr){3-12}

&
& \multirow{3}{*}{Comedy}
& EER
& 7.51 & 9.21 & 2.31 & 9.16
& 9.14 & 3.83 & 3.31 & 5.84 \\

& & & GMR$_{0.01}$
& 4.17 & 40.83 & 73.33 & 26.67
& 27.08 & 68.75 & 91.25 & 48.75 \\

& & & GMR$_{0.1}$
& 26.25 & 68.75 & 86.25 & 41.67
& 69.58 & 72.08 & 94.17 & 68.33 \\
\cmidrule(lr){3-12}

&
& \multirow{3}{*}{Horror}
& EER
& 4.51 & 4.01 & 0.73 & 2.87
& 4.73 & 2.15 & 0.21 & 1.02 \\

& & & GMR$_{0.01}$
& 29.29 & 62.86 & 93.10 & 27.38
& 41.43 & 78.81 & 98.33 & 77.14 \\

& & & GMR$_{0.1}$
& 40.71 & 69.29 & 96.90 & 62.38
& 72.62 & 92.38 & 99.52 & 90.00 \\
\cmidrule(lr){2-12}

& \multirow{9}{*}{Relax}
& \multirow{3}{*}{L7T 1}
& EER
& 3.34 & 1.24 & \textcolor{blue}{0.00} & 1.68
& 2.78 & 1.11 & 0.63 & 1.70 \\

& & & GMR$_{0.01}$
& 55.00 & 66.67 & \textcolor{blue}{100.00} & 73.33
& 61.11 & 80.00 & 96.11 & 71.67 \\

& & & GMR$_{0.1}$
& 72.78 & 90.00 & \textcolor{blue}{100.00} & 83.89
& 67.78 & 97.22 & 98.33 & 83.33 \\
\cmidrule(lr){3-12}

&
& \multirow{3}{*}{Comedy}
& EER
& 5.81 & 7.64 & 1.25 & 7.46
& 4.20 & 4.13 & 2.08 & 5.83 \\

& & & GMR$_{0.01}$
& 3.75 & 38.33 & 82.92 & 42.08
& 40.00 & 70.42 & 91.67 & 59.17 \\

& & & GMR$_{0.1}$
& 34.58 & 76.25 & 92.50 & 62.92
& 58.33 & 76.25 & 94.17 & 72.08 \\
\cmidrule(lr){3-12}

&
& \multirow{3}{*}{Horror}
& EER
& 4.93 & 3.81 & 0.48 & 6.18
& 2.44 & 1.32 & 0.24 & 1.94 \\

& & & GMR$_{0.01}$
& 24.52 & 59.29 & 91.19 & 38.33
& 57.86 & 89.76 & 99.05 & 29.52 \\

& & & GMR$_{0.1}$
& 59.05 & 78.57 & 98.33 & 65.00
& 72.38 & 95.71 & 99.76 & 79.05 \\

\botrule
\end{tabular*}
\vspace{-4mm}
\end{table*}
Condition-level method differences were assessed using the Friedman test, followed by paired Wilcoxon signed-rank tests with Holm--Bonferroni correction ($p<0.05$) \cite{Nguyen2026InclusiveEEG}. Effect sizes were quantified using Kendall's $W$ and matched rank-biserial correlation \cite{Nguyen2026InclusiveEEG,Kerby2014}. Primary participant-aware inference used 2,000 participant-cluster bootstrap replicates, resampling participants with replacement within each cohort while retaining all their probe observations. Common resamples preserved method--condition pairing, with cohorts resampled independently. Bootstrap confidence intervals quantified EER and GMR uncertainty and pairwise contrasts; simultaneous 95\% intervals controlled multiplicity within algorithm, experiment, and cohort comparison families. Experiment-level estimates equally averaged conditions, with inference conditional on fixed galleries and trained models.

\begin{figure}[!htbp]
\vspace{-4mm}
\centering
\includegraphics[width=0.48\textwidth]
{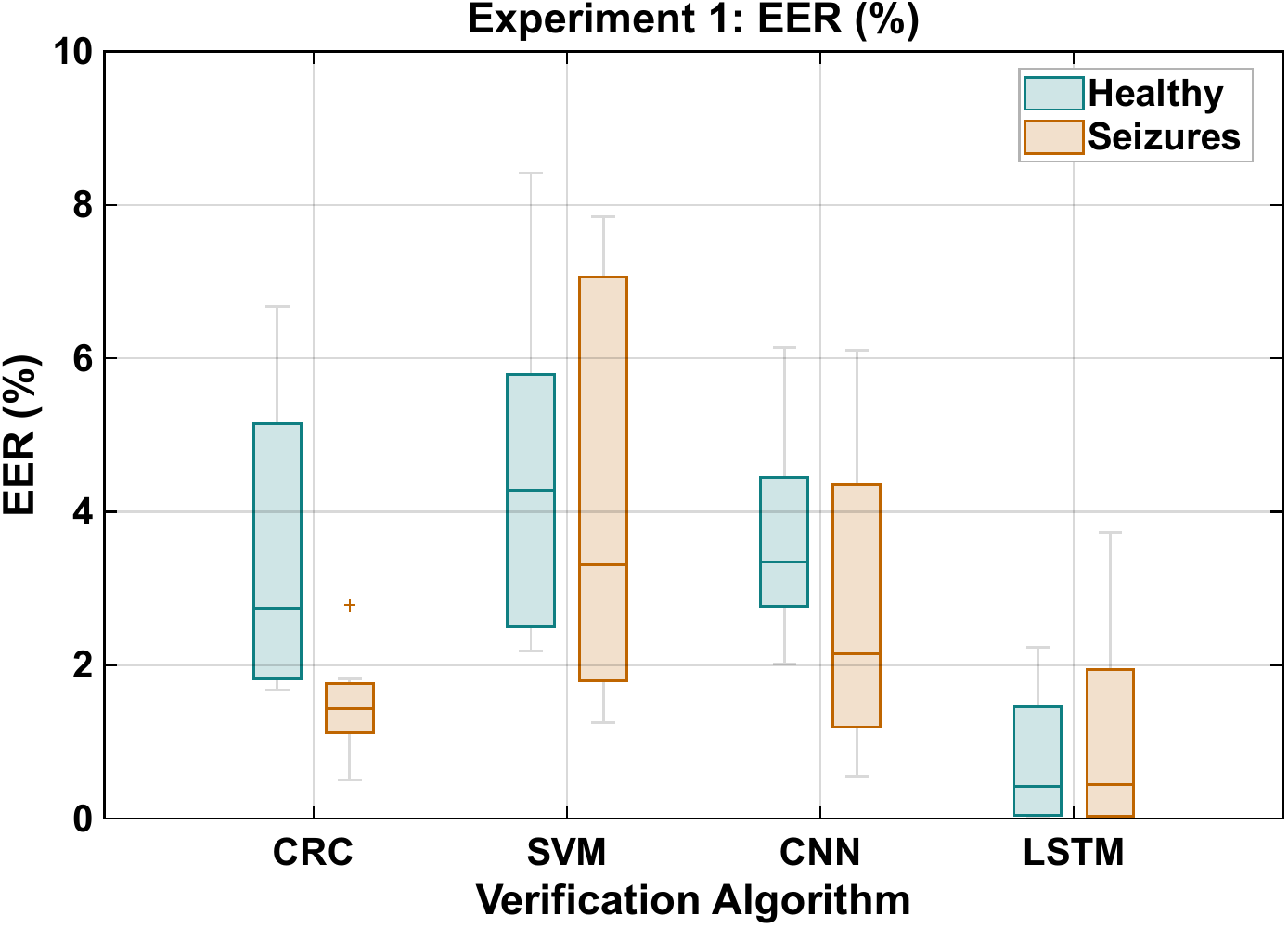}
\hfill
\includegraphics[width=0.48\textwidth]
{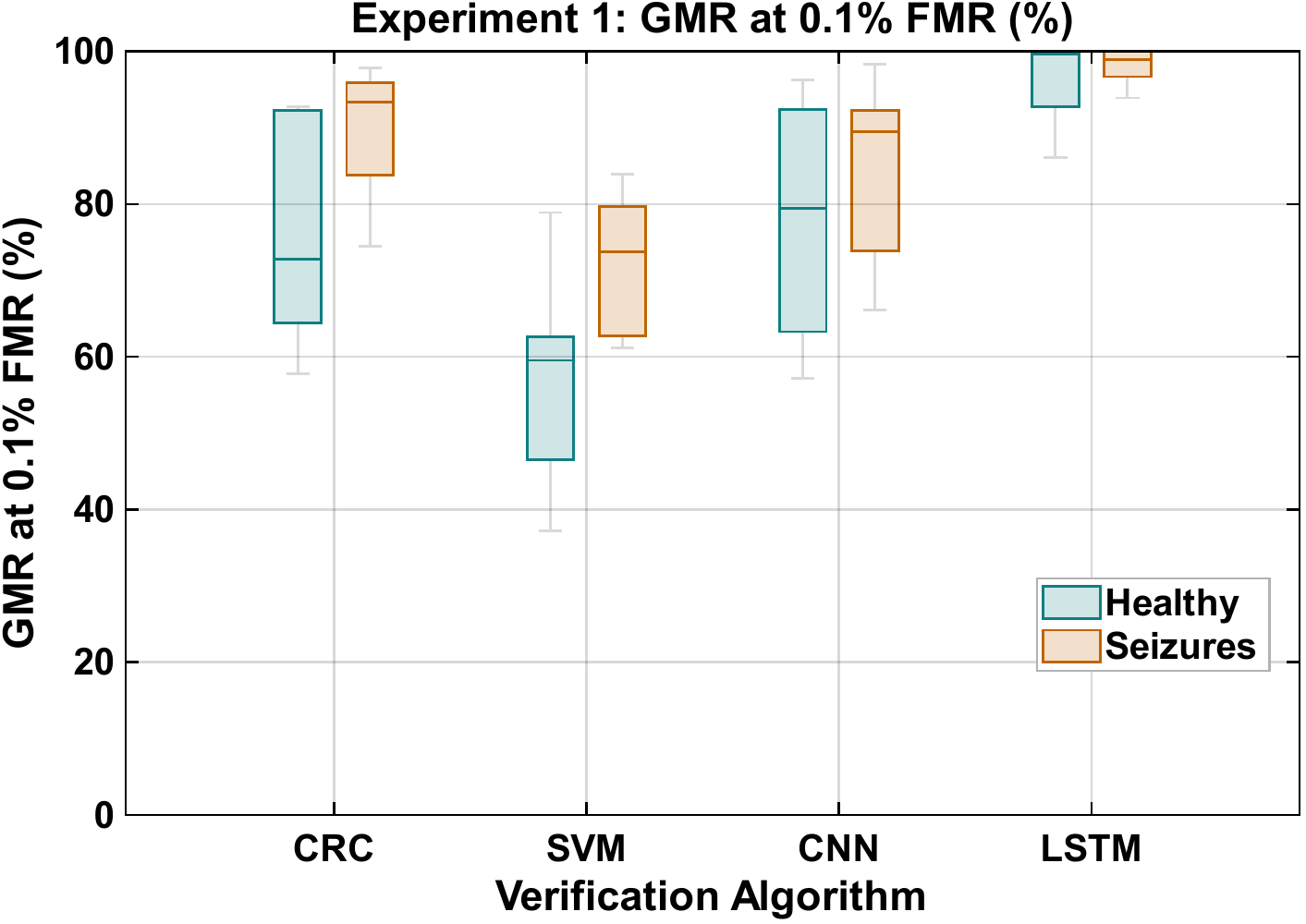}
\vspace{-1mm}
\caption{EER and GMR at \(0.1\%\) FMR for intra-stimulus verification in Experiment~1.}
\label{fig:exp1_eer_gmr}
\vspace{-4mm}
\end{figure}

\begin{figure}[!htbp]
\vspace{-2mm}
\centering
\includegraphics[width=0.48\textwidth]
{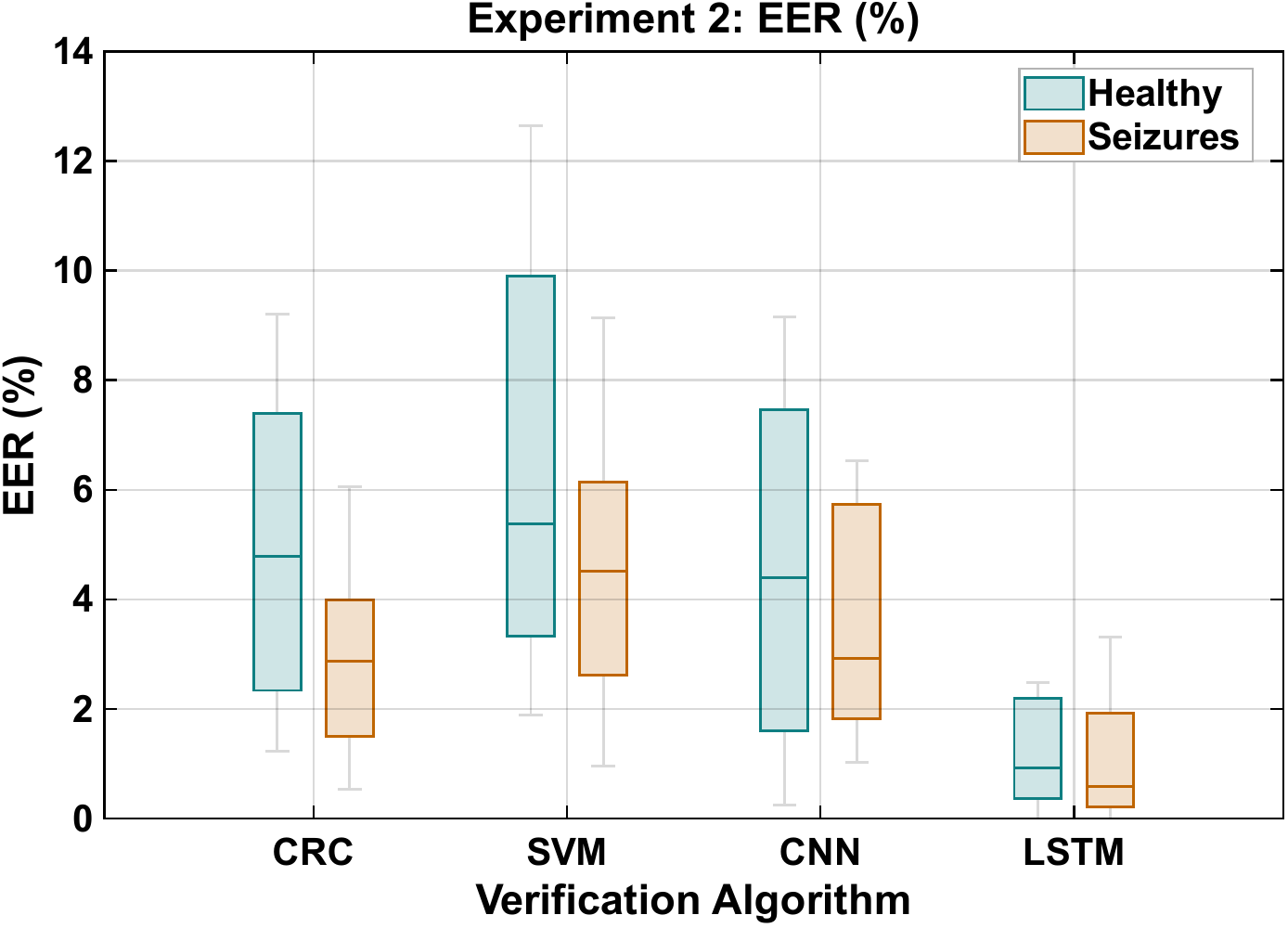}
\hfill
\includegraphics[width=0.48\textwidth]
{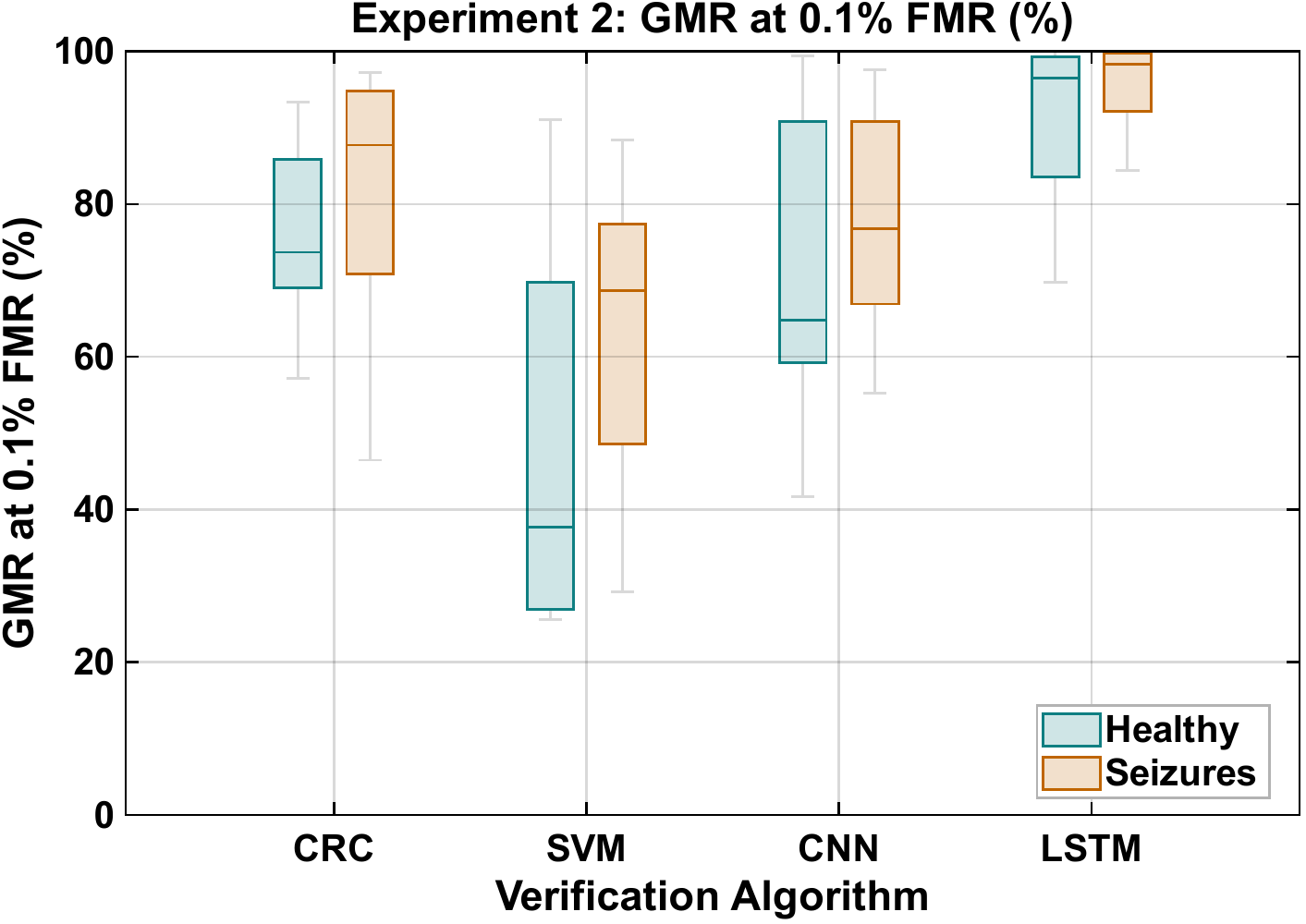}
\vspace{-1mm}
\caption{EER and GMR at \(0.1\%\) FMR for inter-stimulus verification in Experiment~2.}
\label{fig:exp2_eer_gmr}
\vspace{-1mm}
\end{figure}

\begin{figure}[!htbp]
\vspace{-2mm}
\centering
\includegraphics[width=0.48\textwidth]
{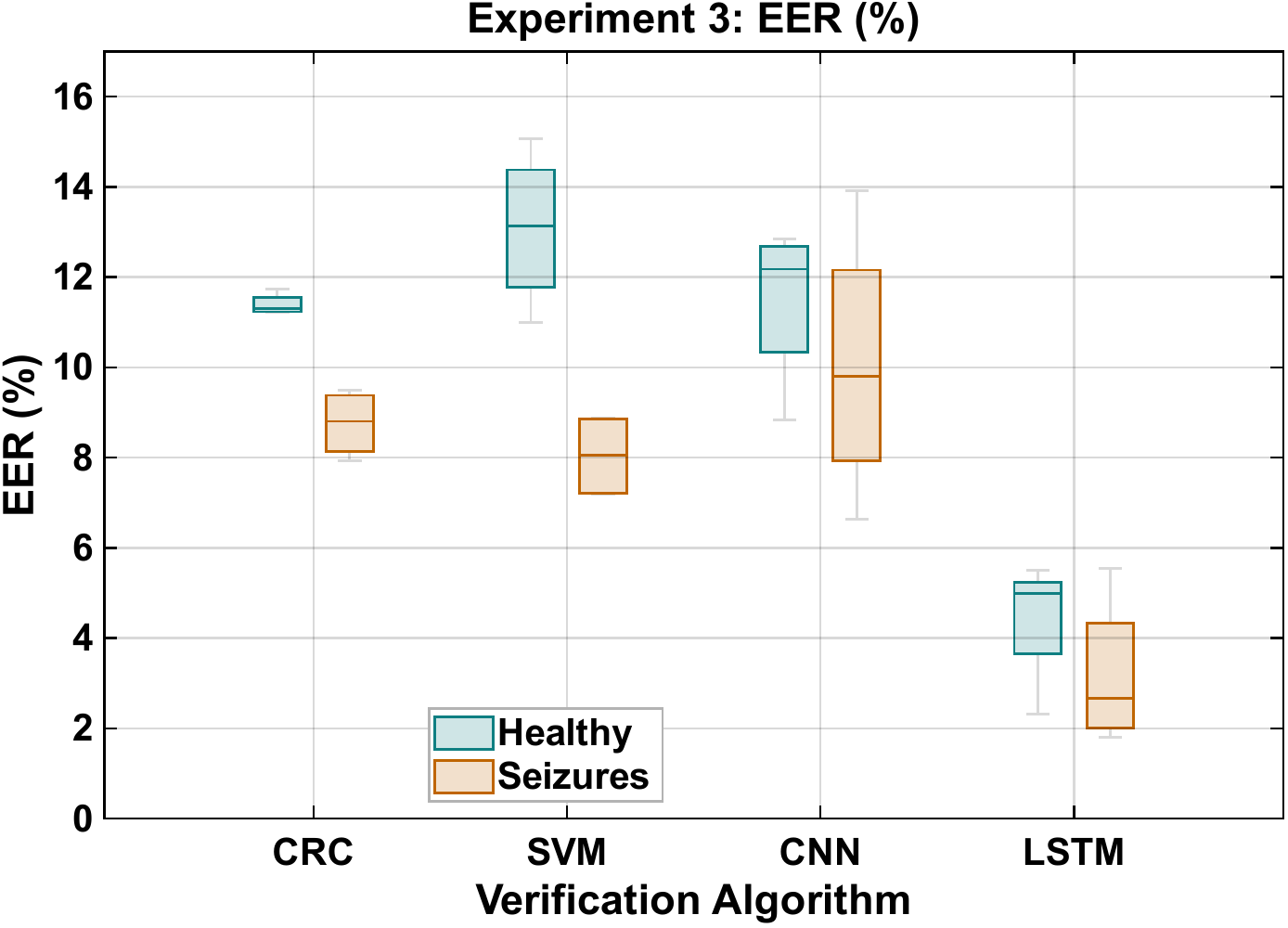}
\hfill
\includegraphics[width=0.48\textwidth]
{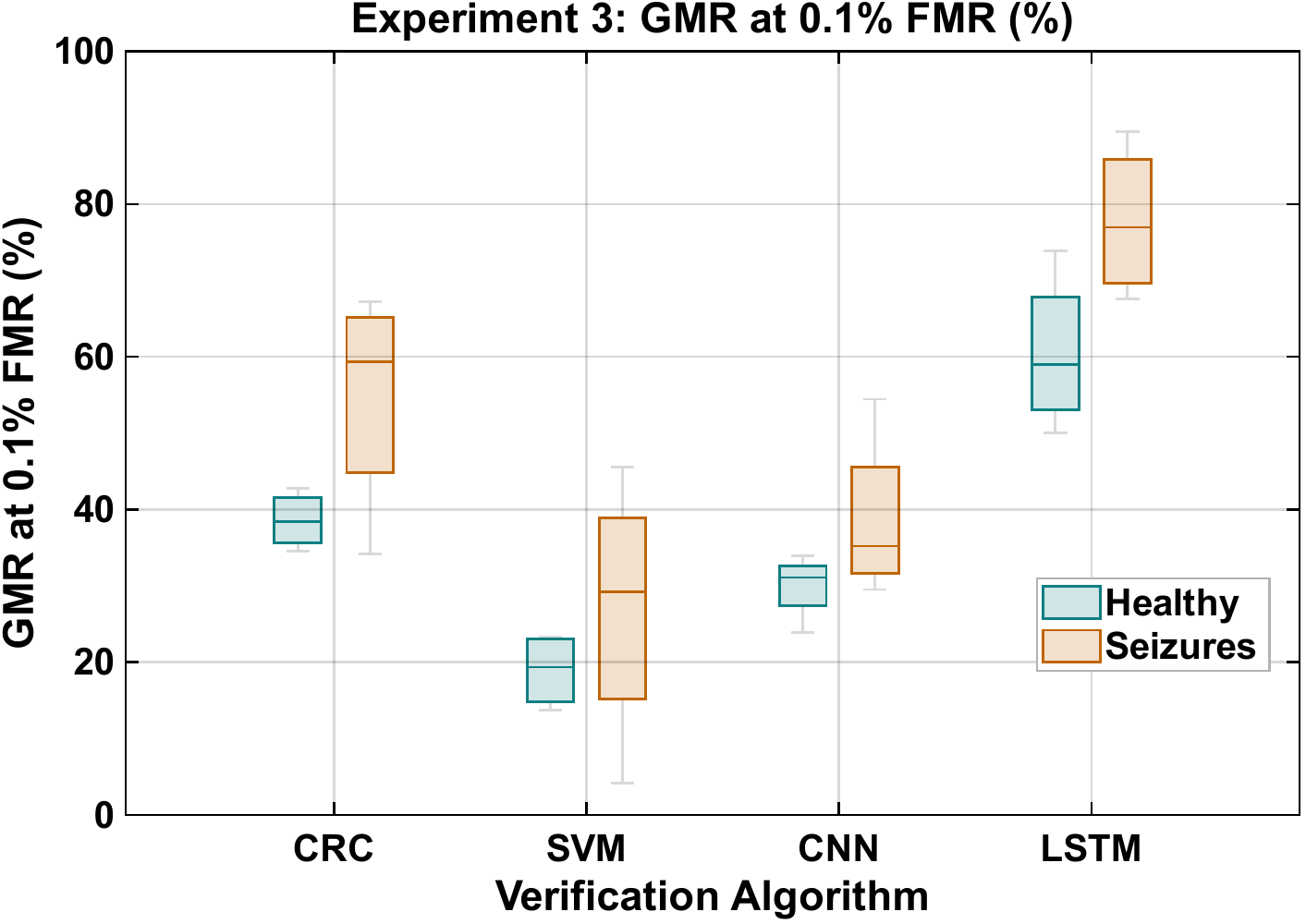}
\vspace{-1mm}
\caption{EER and GMR at \(0.1\%\) FMR for cross-DCP inter-stimulus
verification in Experiment~3.}
\label{fig:exp3_eer_gmr}
\vspace{-4mm}
\end{figure}

The findings are presented in order of increasing enrolment--probe mismatch, beginning with the matched-stimulus baseline and progressing to within-DCP and cross-DCP stimulus comparisons. Figures~\ref{fig:exp1_eer_gmr}--\ref{fig:exp3_eer_gmr} summarize the EER and GMR distributions at $0.1\%$ FMR, with corresponding GMR plots at 0.01\% FMR provided in Supplementary Figs.~S1--S3, while Supplementary Table~S1 reports intra-stimulus results, and Tables~\ref{tab:experiment2_results} and~\ref{tab:experiment3_results} provide within-DCP and cross-DCP results, respectively, including GMR at $0.01\%$ and $0.1\%$ FMR. Secondary participant-level closed-set identification averaged classifier identity scores across each participant's probe segments and selected the highest-scoring identity. Correctly identified participants (out of 30 per cohort) are reported for Experiments 1--3 in Supplementary Tables S2--S4, respectively, separately from the primary verification metrics (EER and GMR).

\textbf{Experiment~1} established the matched-stimulus baseline. As shown in Fig.~\ref{fig:exp1_eer_gmr}, BiLSTM exhibited the lowest median EER and highest median GMR in both cohorts. CRC and CNN generally occupied intermediate positions, and SVM showed the least favourable and most variable performance. At the condition level, all six Friedman comparisons remained significant after Holm correction; however, no pairwise algorithm comparison remained significant under that analysis. The algorithm ordering should therefore be interpreted as a consistent trend rather than conclusive pairwise superiority. Resting-state and video repetitions generally supported reliable verification, while the L7T performance often weakened as the temporal separation increased, suggesting within-session variation even under the same stimulus. Participant-cluster bootstrap analysis did not establish a consistent healthy--epilepsy performance difference.
 
\FloatBarrier

\textbf{Experiment~2} introduced stimulus mismatch within the same DCP, generally shifting EER upward and GMR downward relative to Experiment~1 as shown in Figs.~\ref{fig:exp1_eer_gmr} and~\ref{fig:exp2_eer_gmr}. BiLSTM remained the most
robust method and showed significant condition-level improvements over CRC, SVM, and CNN for EER and both GMR operating points in both cohorts (Table~\ref{tab:experiment2_results}). CRC was the strongest conventional classifier, CNN showed intermediate variation, and SVM was more sensitive to stimulus mismatch. Performance was also directional because reversing enrolment and probe stimuli did not always produce equivalent results. Relaxation-based enrolment transferred favourably to several probe conditions, although no stimulus was universally optimal. Participant-cluster bootstrap analysis did not establish a consistent healthy--epilepsy performance difference in this experiment.

\FloatBarrier

\vspace{-6mm}

\textbf{Experiment~3} introduced the greatest mismatch by comparing different stimuli across DCP-1 and DCP-2. It produced descriptively higher EER and lower GMR than the first two experiments as shown in Figs.~\ref{fig:exp3_eer_gmr}, confirming cross-DCP verification as the most challenging setting. BiLSTM retained the best median performance (Table~\ref{tab:experiment3_results}). After Holm correction, the condition-level algorithm effect remained significant for both GMR measures in both cohorts, but not for EER, and no post-hoc algorithm comparison remained significant. CRC generally achieved the strongest GMR among the conventional methods, whereas SVM and CNN were more condition dependent. L7T and relaxation were frequently favourable probe conditions, particularly for BiLSTM. Across the 12 algorithm--metric combinations (four algorithms $\times$ three metrics), the epilepsy cohort showed more favourable descriptive performance in 11 combinations; however, participant-cluster bootstrap analysis did not establish a consistent cohort difference. The larger cross-DCP performance decline suggests that enrolment--probe mismatch was a more prominent descriptive pattern than cohort status.

\begin{table*}[t]
\caption{Experiment 3: cross-protocol verification using enrolment
data from DCP-1 and probe data from DCP-2. All values are in \%; GMR
subscripts denote the corresponding FMR operating points.}
\label{tab:experiment3_results}%
\centering
\scriptsize
\setlength{\tabcolsep}{1.5pt}
\renewcommand{\arraystretch}{1.02}

\begin{tabular*}{\textwidth}
{@{\extracolsep\fill}lll*{8}{c}@{}}
\toprule
\multirow{2}{*}{\shortstack{Enrolment\\(DCP-1)}} &
\multirow{2}{*}{\shortstack{Probe\\(DCP-2)}} &
\multirow{2}{*}{Metric} &
\multicolumn{4}{c}{Healthy cohort} &
\multicolumn{4}{c}{Epilepsy cohort} \\
\cmidrule(lr){4-7}\cmidrule(lr){8-11}
& & &
SVM & CRC & BiLSTM & CNN &
SVM & CRC & BiLSTM & CNN \\
\midrule

\multirow{12}{*}{REC 1,2}
& \multirow{3}{*}{L7T 1}
& EER
& 11.00 & 11.22 & \textcolor{blue}{2.31} & 12.84
& 8.87 & 9.49 & \textcolor{blue}{1.79} & 9.21 \\

& & GMR$_{0.01}$
& 14.44 & 23.89 & \textcolor{blue}{70.00} & 17.22
& 8.33 & 22.78 & \textcolor{blue}{72.78} & 7.22 \\

& & GMR$_{0.1}$
& 23.33 & 36.67 & \textcolor{blue}{73.89} & 23.89
& 32.22 & 55.56 & \textcolor{blue}{82.22} & 36.67 \\
\cmidrule(lr){2-11}

& \multirow{3}{*}{Relax}
& EER
& 15.07 & 11.73 & 4.98 & 8.83
& 7.19 & 8.34 & 2.21 & 10.40 \\

& & GMR$_{0.01}$
& 12.78 & 33.33 & 39.44 & 16.11
& 28.33 & 37.22 & 79.44 & 27.78 \\

& & GMR$_{0.1}$
& 22.78 & 42.78 & 50.00 & 33.89
& 45.56 & 67.22 & 89.44 & 54.44 \\
\cmidrule(lr){2-11}

& \multirow{3}{*}{Comedy}
& EER
& 12.57 & 11.24 & 4.99 & 11.82
& 8.86 & 7.94 & 5.54 & 6.64 \\

& & GMR$_{0.01}$
& 6.67 & 16.67 & 42.08 & 11.67
& 1.25 & 21.25 & 39.58 & 5.83 \\

& & GMR$_{0.1}$
& 13.75 & 34.58 & 61.67 & 31.25
& 4.17 & 34.17 & 71.67 & 33.75 \\
\cmidrule(lr){2-11}

& \multirow{3}{*}{Horror}
& EER
& 13.69 & 11.35 & 5.50 & 12.53
& 7.24 & 9.27 & 3.11 & 13.91 \\

& & GMR$_{0.01}$
& 5.24 & 15.48 & 34.76 & 8.81
& 14.29 & 23.33 & 56.67 & 14.76 \\

& & GMR$_{0.1}$
& 15.95 & 40.24 & 56.19 & 30.95
& 26.19 & 63.10 & 67.62 & 29.52 \\

\botrule
\end{tabular*}
\vspace{-5mm}
\end{table*}

Across all three experiments, EER generally increased and GMR decreased with increasing enrolment--probe mismatch. BiLSTM achieved the best median performance in all 18 experiment--cohort--metric combinations (three experiments $\times$ two cohorts $\times$ three metrics), CRC was the strongest conventional method, and SVM was generally the most sensitive to stimulus and DCP variation under fixed hyperparameters, without gallery-based tuning. At the condition level, 16 of the 18 Friedman tests remained significant after Holm correction, with large Kendall's $W$ values of $0.657$--$1.000$; the exceptions were the Experiment~3 EER comparisons for both cohorts. Corrected pairwise algorithm differences occurred only in Experiment~2. Participant-cluster bootstrap analysis further supported algorithm differences in 60 of 276 EER contrasts using simultaneous 95\% confidence intervals, including 50 favouring BiLSTM. Experiment-level bootstrap comparisons showed higher EER in Experiment~3 than in Experiments~1 and~2 in 16 of 24 contrasts (simultaneous 95\% CIs).
Cohort differences were less consistent, with only one of 12 experiment-level EER contrasts excluding zero under simultaneous intervals. Thus, no uniform epilepsy-related
performance reduction was established.

Taken together, the absence of a consistent performance difference between the two cohorts suggests that the identity-related EEG characteristics remained sufficiently repeatable and discriminative in the medicated epilepsy cohort under the evaluated conditions, although equivalence in verification performance between cohorts cannot be inferred. All participants with epilepsy were receiving prescribed antiseizure medication, which is known to modify quantitative EEG characteristics \cite{Dini2023ASMEEG}; medication-related modulation may therefore have contributed to the broadly comparable verification performance, although its specific effect cannot be determined from the present study. Importantly, this result does not imply an absence of epilepsy-related EEG alterations, as previous clinical investigations have reported interictal changes in spectral activity, synchronization, and functional-network organization \cite{Pyrzowski2015Focal,Pegg2020Interictal,Woldman2020Networks}. 
Accordingly, the present findings provide a baseline assessment of EEG-based verification in medicated participants with epilepsy, limited to a single-centre cohort without external validation. Future research should extend this evaluation to larger cohorts and examine medication-naive participants or clinically supervised off-medication recordings, where medically and ethically permissible, to determine whether the absence of antiseizure medication alters the repeatability, discriminability, and resulting verification performance of subject-specific EEG characteristics, while also assessing computational efficiency, real-time feasibility, and practical EEG acquisition requirements for deployment.

\vspace{-3mm}
\section{Conclusion}
\label{sec:conclusion}
EEG-based biometric verification uses internally generated, subject-specific brain patterns, yet its reliability in participants with epilepsy remains largely unexplored. Addressing this gap, this baseline study systematically evaluated verification using a balanced EEG database of
60 participants---30 healthy and 30 with epilepsy--- comprising 207,360 channel-wise 10-s segments acquired under multiple stimulation conditions and two DCPs. To examine the influence of epilepsy-related spectral alterations, log-absolute power features from five canonical EEG bands were evaluated using CRC, SVM, CNN, and BiLSTM under progressively increasing enrolment--probe mismatch. Performance declined from matched-stimulus to within-DCP and cross-DCP verification. Across the six experiment--cohort combinations, BiLSTM achieved mean EERs of
$0.78$--$4.44\%$ and mean GMRs of $60.44$--$98.17\%$ at $0.1\%$ FMR, whereas CRC, the strongest conventional method, achieved corresponding ranges of $1.48$--$11.38\%$ and $38.57$--$89.95\%$. At the condition level, 16 of 18 Friedman comparisons remained significant after
Holm correction. Participant-cluster bootstrap analysis further supported algorithm differences
and higher cross-DCP EER, while no consistent healthy--epilepsy performance difference was
established. These findings establish a verification baseline under medicated, interictal conditions and motivate further evaluation using larger cohorts and, where clinically and ethically permissible, medication-naive or medically supervised off-medication recordings.

\backmatter

\begingroup
\setlength{\bibsep}{2.8pt}
\bibliography{sn-bibliography}
\endgroup

\end{document}


\renewcommand{\thetable}{S\arabic{table}}
\renewcommand{\thefigure}{S\arabic{figure}}

\begin{center}
{\Large\bfseries Supplementary Material}

\vspace{3pt}

{\normalsize EEG-Based Biometric Verification in Epilepsy:
A Baseline Evaluation Across Multiple Stimulus Conditions}

\vspace{3pt}

{\small CVIP 2026 --- Paper ID 875}
\end{center}

\vspace{5pt}

\noindent\textbf{Supplementary verification results:
Experiment 1 (intra-stimulus verification).}

\begin{table}[!htbp]
\caption{Experiment 1: intra-stimulus verification within each
 data-collection protocol. All values are in \%; GMR subscripts denote
 the corresponding FMR operating points.}
\label{tab:S1_verification}%
\centering
\scriptsize
\setlength{\tabcolsep}{1.5pt}
\renewcommand{\arraystretch}{1.02}
\begin{tabular*}{\textwidth}
{@{\extracolsep\fill}llll*{8}{c}@{}}
\toprule
\multirow{2}{*}{DCP} &
\multirow{2}{*}{Enrolment} &
\multirow{2}{*}{Probe} &
\multirow{2}{*}{Metric} &
\multicolumn{4}{c}{Healthy subjects} &
\multicolumn{4}{c}{Seizure subjects} \\
\cmidrule(lr){5-8}\cmidrule(lr){9-12}
& & & &
SVM & CRC & LSTM & CNN &
SVM & CRC & LSTM & CNN \\
\midrule
\multirow{12}{*}{DCP-1}
& \multirow{3}{*}{REC1} & \multirow{3}{*}{REC2}
& EER
& 2.21 & 1.68 & 0.05 & 2.77 & 1.25 & 0.50 & 0.00 & 1.11 \\

& &
& GMR$_{0.01}$
& 63.89 & 74.44 & 99.44 & 72.78 & 60.00 & 92.22 & 100.00 & 81.11 \\

& &
& GMR$_{0.1}$
& 78.89 & 91.67 & 100.00 & 93.33 & 83.89 & 97.78 & 100.00 & 91.67 \\
\cmidrule(lr){2-12}

& \multirow{3}{*}{Relax1} & \multirow{3}{*}{Relax2}
& EER
& 8.42 & 5.56 & 1.58 & 3.34 & 3.31 & 1.11 & 1.11 & 2.67 \\

& &
& GMR$_{0.01}$
& 33.33 & 37.22 & 83.33 & 76.11 & 17.78 & 91.11 & 98.33 & 59.44 \\

& &
& GMR$_{0.1}$
& 52.78 & 57.78 & 86.11 & 89.44 & 62.22 & 96.11 & 98.89 & 89.44 \\
\cmidrule(lr){2-12}

& \multirow{3}{*}{Comedy1} & \multirow{3}{*}{Comedy2}
& EER
& 4.28 & 2.74 & 0.04 & 4.79 & 2.22 & 1.43 & 0.03 & 1.43 \\

& &
& GMR$_{0.01}$
& 43.33 & 48.57 & 99.05 & 61.90 & 75.71 & 60.48 & 99.05 & 84.29 \\

& &
& GMR$_{0.1}$
& 59.52 & 67.62 & 100.00 & 71.43 & 81.43 & 91.90 & 100.00 & 92.38 \\
\cmidrule(lr){2-12}

& \multirow{3}{*}{Horror1} & \multirow{3}{*}{Horror2}
& EER
& 2.18 & 2.08 & 0.42 & 2.02 & 1.66 & 1.60 & 0.45 & 2.15 \\

& &
& GMR$_{0.01}$
& 10.42 & 80.42 & 97.08 & 92.50 & 53.75 & 87.08 & 90.00 & 50.83 \\

& &
& GMR$_{0.1}$
& 62.92 & 92.50 & 99.58 & 96.25 & 73.75 & 93.33 & 98.33 & 80.42 \\
\midrule
\multirow{9}{*}{DCP-2}
& \multirow{3}{*}{L7T 1} & \multirow{3}{*}{L7T 2}
& EER
& 3.34 & 1.73 & 0.01 & 2.77 & 5.00 & 1.14 & 0.03 & 0.56 \\

& &
& GMR$_{0.01}$
& 48.33 & 79.44 & 99.44 & 71.11 & 53.89 & 89.44 & 98.33 & 91.67 \\

& &
& GMR$_{0.1}$
& 61.67 & 92.78 & 100.00 & 79.44 & 74.44 & 95.00 & 100.00 & 98.33 \\
\cmidrule(lr){2-12}

& \multirow{3}{*}{L7T 1} & \multirow{3}{*}{L7T 3}
& EER
& 5.00 & 3.92 & 1.11 & 3.41 & 7.75 & 2.78 & 2.22 & 6.10 \\

& &
& GMR$_{0.01}$
& 16.67 & 58.33 & 85.56 & 38.89 & 45.56 & 66.11 & 89.44 & 12.78 \\

& &
& GMR$_{0.1}$
& 37.22 & 63.33 & 94.44 & 57.22 & 64.44 & 74.44 & 93.89 & 66.11 \\
\cmidrule(lr){2-12}

& \multirow{3}{*}{L7T 1} & \multirow{3}{*}{L7T 4}
& EER
& 6.05 & 6.68 & 2.23 & 6.14 & 7.84 & 1.82 & 3.74 & 4.91 \\

& &
& GMR$_{0.01}$
& 13.89 & 48.33 & 84.44 & 36.67 & 33.89 & 71.11 & 93.33 & 19.44 \\

& &
& GMR$_{0.1}$
& 44.44 & 72.78 & 92.22 & 60.56 & 61.11 & 81.11 & 96.11 & 71.67 \\
\bottomrule
\end{tabular*}
\end{table}

\noindent\textbf{Supplementary analysis:
participant-level closed-set identification.}

This exploratory analysis was performed using the
saved identity-score matrices from four classifiers
(SVM, CRC, CNN, and BiLSTM). Within each cohort and
stimulus comparison, the gallery contained 30 enrolled
identities. Probe scores were grouped by the participant
to whom the probe segments belonged.

For each participant, the scores for each candidate
identity were averaged across that participant's probe
segments. The candidate with the highest mean score
was assigned as the predicted identity. The number of
participants whose predicted identity matched their
true identity was counted out of 30.


\begin{table}[!htbp]
\centering
\caption{Experiment 1: intra-stimulus closed-set participant identification. Correctly identified participants out of 30.}
\label{tab:S1_identification}
\setlength{\tabcolsep}{4.3pt}
\renewcommand{\arraystretch}{1.13}
\scriptsize
\begin{tabular}{@{}lc*{8}{c}@{}}
\toprule
& & \multicolumn{4}{c}{Healthy ($n=30$)} & \multicolumn{4}{c}{Epilepsy ($n=30$)} \\
\cmidrule(lr){3-6}\cmidrule(l){7-10}
Gallery $\rightarrow$ probe & DCP & SVM & CRC & CNN & BiLSTM & SVM & CRC & CNN & BiLSTM \\
\midrule
REC1 $\rightarrow$ REC2 & DCP-1 &
30/30 & 30/30 & 29/30 & 30/30 &
30/30 & 30/30 & 30/30 & 30/30 \\
Relax1 $\rightarrow$ Relax2 & DCP-1 &
28/30 & 28/30 & 29/30 & 28/30 &
29/30 & 30/30 & 30/30 & 30/30 \\
Comedy1 $\rightarrow$ Comedy2 & DCP-1 &
29/30 & 29/30 & 28/30 & 30/30 &
30/30 & 30/30 & 30/30 & 30/30 \\
Horror1 $\rightarrow$ Horror2 & DCP-1 &
30/30 & 30/30 & 30/30 & 30/30 &
30/30 & 30/30 & 28/30 & 30/30 \\
\midrule
L7Tone $\rightarrow$ L7Ttwo & DCP-2 &
29/30 & 30/30 & 28/30 & 30/30 &
28/30 & 29/30 & 30/30 & 30/30 \\
L7Tone $\rightarrow$ L7Tthree & DCP-2 &
29/30 & 29/30 & 29/30 & 29/30 &
28/30 & 29/30 & 27/30 & 29/30 \\
L7Tone $\rightarrow$ L7Tfour & DCP-2 &
28/30 & 29/30 & 28/30 & 29/30 &
29/30 & 30/30 & 28/30 & 29/30 \\
\bottomrule
\end{tabular}
\end{table}

\begin{table}[!htbp]
\centering
\caption{Experiment 2: within-DCP inter-stimulus closed-set participant identification. Correctly identified participants out of 30.}
\label{tab:S2_identification}
\setlength{\tabcolsep}{4.3pt}
\renewcommand{\arraystretch}{1.13}
\scriptsize
\begin{tabular}{@{}lc*{8}{c}@{}}
\toprule
& & \multicolumn{4}{c}{Healthy ($n=30$)} & \multicolumn{4}{c}{Epilepsy ($n=30$)} \\
\cmidrule(lr){3-6}\cmidrule(l){7-10}
Gallery $\rightarrow$ probe & DCP & SVM & CRC & CNN & BiLSTM & SVM & CRC & CNN & BiLSTM \\
\midrule
REC12 $\rightarrow$ Relax12 & DCP-1 &
24/30 & 29/30 & 27/30 & 29/30 &
28/30 & 29/30 & 27/30 & 30/30 \\
REC12 $\rightarrow$ Comedy12 & DCP-1 &
23/30 & 28/30 & 28/30 & 27/30 &
27/30 & 29/30 & 26/30 & 29/30 \\
REC12 $\rightarrow$ Horror12 & DCP-1 &
20/30 & 28/30 & 27/30 & 27/30 &
28/30 & 29/30 & 26/30 & 29/30 \\
Relax12 $\rightarrow$ REC12 & DCP-1 &
24/30 & 28/30 & 29/30 & 29/30 &
28/30 & 30/30 & 28/30 & 29/30 \\
Relax12 $\rightarrow$ Comedy12 & DCP-1 &
30/30 & 30/30 & 30/30 & 30/30 &
30/30 & 30/30 & 30/30 & 30/30 \\
Relax12 $\rightarrow$ Horror12 & DCP-1 &
30/30 & 30/30 & 30/30 & 30/30 &
30/30 & 30/30 & 30/30 & 30/30 \\
\midrule
L7Tone $\rightarrow$ Relax & DCP-2 &
30/30 & 30/30 & 30/30 & 30/30 &
30/30 & 30/30 & 30/30 & 30/30 \\
L7Tone $\rightarrow$ Comedy & DCP-2 &
29/30 & 27/30 & 26/30 & 29/30 &
28/30 & 28/30 & 26/30 & 29/30 \\
L7Tone $\rightarrow$ Horror & DCP-2 &
28/30 & 29/30 & 28/30 & 30/30 &
28/30 & 30/30 & 30/30 & 30/30 \\
\midrule
Relax $\rightarrow$ L7Tone & DCP-2 &
29/30 & 30/30 & 30/30 & 30/30 &
30/30 & 30/30 & 29/30 & 30/30 \\
Relax $\rightarrow$ Comedy & DCP-2 &
29/30 & 29/30 & 27/30 & 29/30 &
28/30 & 29/30 & 27/30 & 29/30 \\
Relax $\rightarrow$ Horror & DCP-2 &
30/30 & 30/30 & 27/30 & 30/30 &
30/30 & 30/30 & 29/30 & 30/30 \\
\bottomrule
\end{tabular}
\end{table}

\begin{table}[!htbp]
\centering
\caption{Experiment 3: cross-DCP inter-stimulus closed-set participant identification (DCP-1 gallery, DCP-2 probe). Correctly identified participants out of 30.}
\label{tab:S3_identification}
\setlength{\tabcolsep}{4.3pt}
\renewcommand{\arraystretch}{1.13}
\scriptsize
\begin{tabular}{@{}lc*{8}{c}@{}}
\toprule
& & \multicolumn{4}{c}{Healthy ($n=30$)} & \multicolumn{4}{c}{Epilepsy ($n=30$)} \\
\cmidrule(lr){3-6}\cmidrule(l){7-10}
Gallery $\rightarrow$ probe & DCP & SVM & CRC & CNN & BiLSTM & SVM & CRC & CNN & BiLSTM \\
\midrule
REC12 $\rightarrow$ L7Tone & P1$\rightarrow$P2 &
20/30 & 23/30 & 19/30 & 28/30 &
26/30 & 24/30 & 24/30 & 28/30 \\
REC12 $\rightarrow$ Relax & P1$\rightarrow$P2 &
22/30 & 23/30 & 22/30 & 26/30 &
27/30 & 26/30 & 23/30 & 29/30 \\
REC12 $\rightarrow$ Comedy & P1$\rightarrow$P2 &
20/30 & 23/30 & 23/30 & 27/30 &
23/30 & 25/30 & 24/30 & 26/30 \\
REC12 $\rightarrow$ Horror & P1$\rightarrow$P2 &
21/30 & 27/30 & 21/30 & 24/30 &
25/30 & 27/30 & 22/30 & 25/30 \\
\bottomrule
\end{tabular}
\end{table}

\noindent\textit{Notes.} DCP-1 and DCP-2 denote data-collection protocols 1 and 2. REC denotes resting eyes-closed EEG; 7LT denotes alphanumeric flicker stimulation. CRC = collaborative representation classifier; CNN = convolutional neural network; BiLSTM = bidirectional long short-term memory. For DCP-1, the suffix ``12'' denotes the combined recordings of the same stimulus. Each entry is the number of correctly identified participants out of 30 in the corresponding cohort, using mean-score aggregation over probe segments. These exploratory closed-set identification results are distinct from the verification metrics (EER and GMR) reported in the main manuscript; they do not demonstrate generalization to unseen identities or recording sessions.

\section*{Supplementary Verification Analysis: GMR at 0.01\% FMR}

\begin{figure}[!htbp]
\centering
\includegraphics[width=0.60\textwidth]{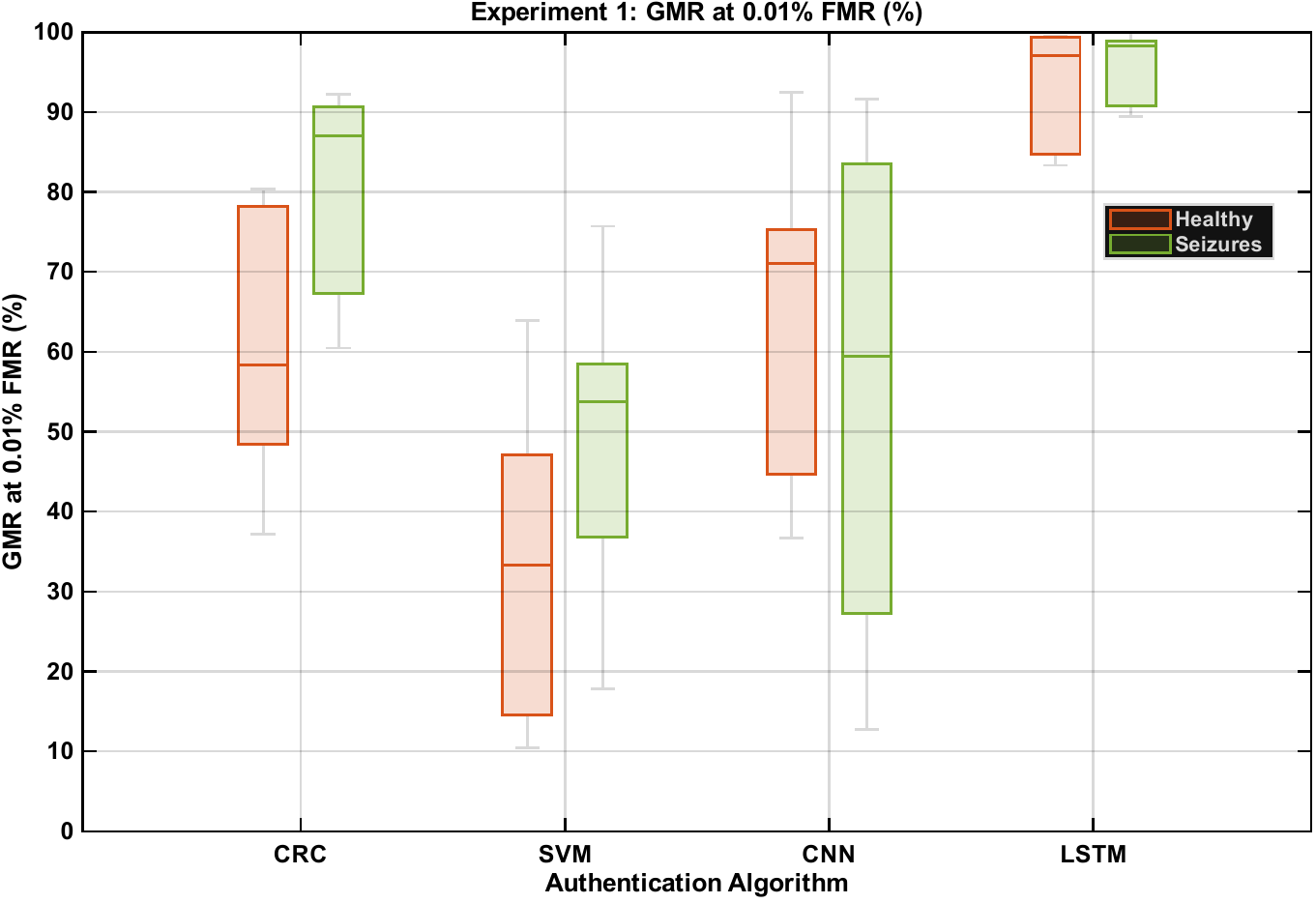}
\caption{Experiment 1: GMR at 0.01\% FMR for intra-stimulus verification in the healthy and epilepsy cohorts.}
\label{fig:S1_gmr}
\end{figure}

\begin{figure}[!htbp]
\centering
\includegraphics[width=0.60\textwidth]{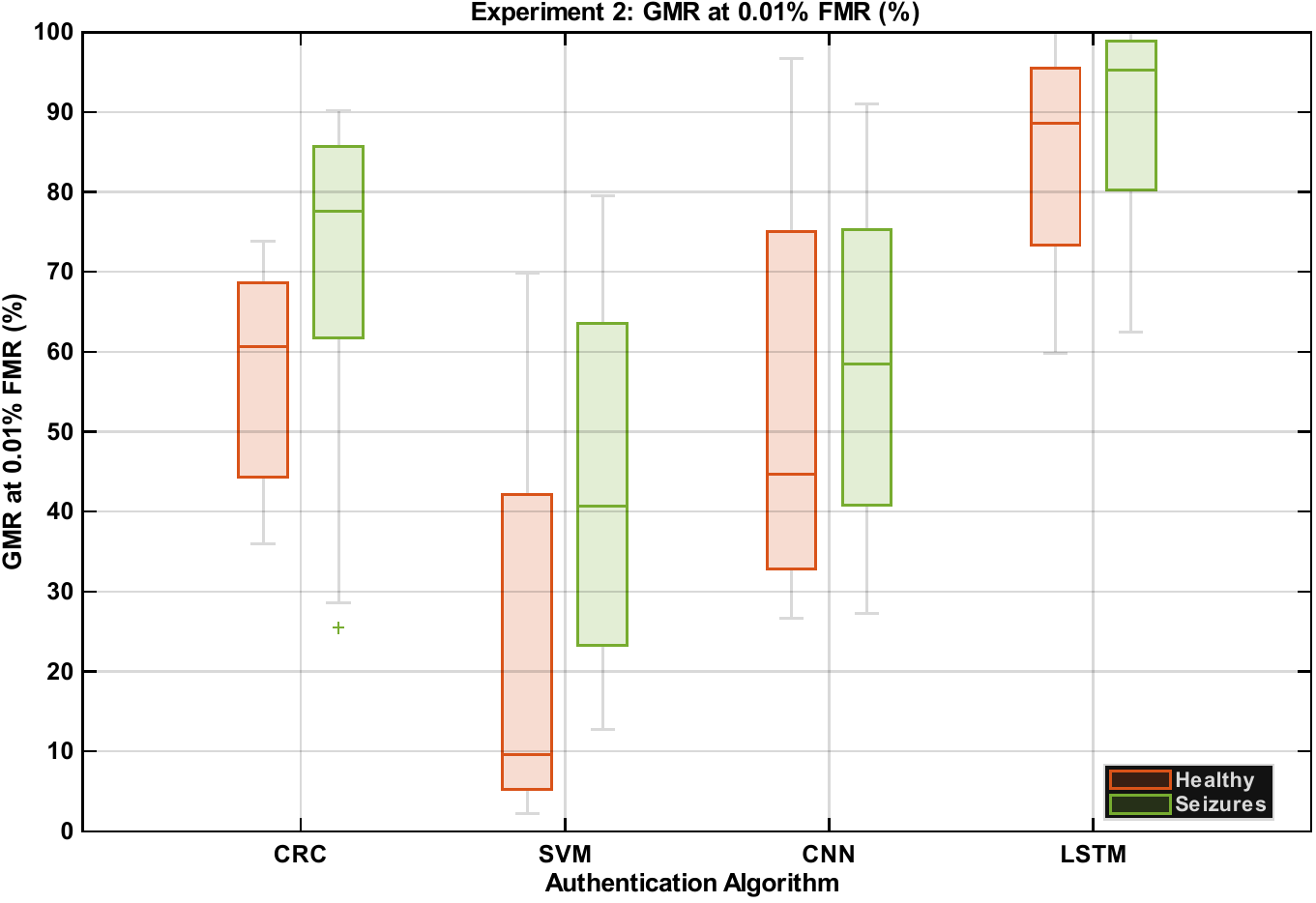}
\caption{Experiment 2: GMR at 0.01\% FMR for within-DCP inter-stimulus verification in the healthy and epilepsy cohorts.}
\label{fig:S2_gmr}
\end{figure}

\begin{figure}[!htbp]
\centering
\includegraphics[width=0.60\textwidth]{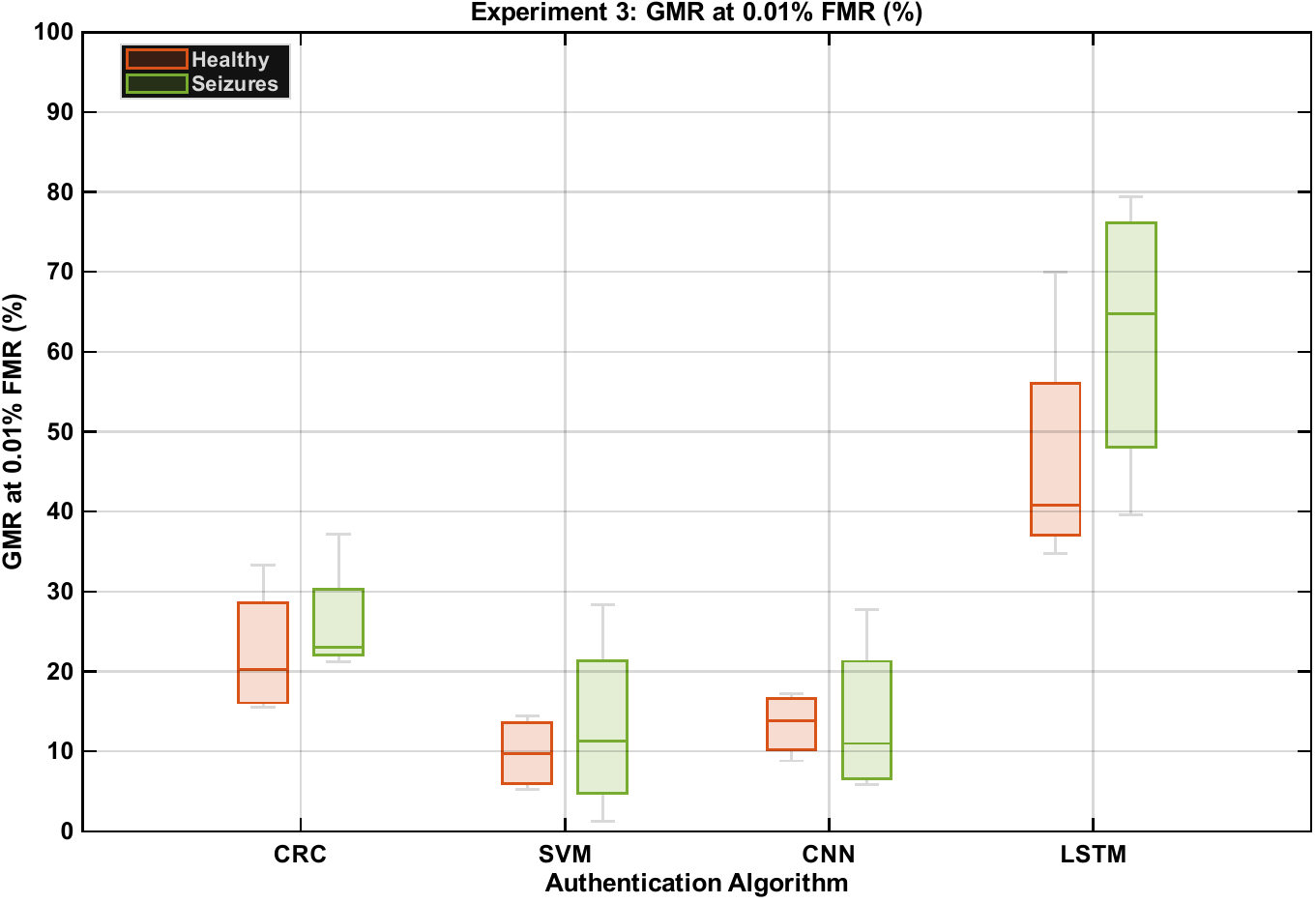}
\caption{Experiment 3: GMR at 0.01\% FMR for cross-DCP inter-stimulus verification in the healthy and epilepsy cohorts.}
\label{fig:S3_gmr}
\end{figure}

Figures~\ref{fig:S1_gmr}--\ref{fig:S3_gmr} present
the genuine match rate (GMR) distributions at a false
match rate (FMR) of 0.01\% for Experiments 1--3,
respectively, complementing the corresponding results
at 0.1\% FMR reported in Figures 3--5 of the main
manuscript. The lower FMR threshold represents a more
stringent verification operating point, permitting
fewer false matches. For the same verification-score
evaluation, GMR at 0.01\% FMR is therefore expected
to be less than or equal to GMR at 0.1\% FMR.

This comparison provides an additional perspective
on verification performance across the three
experimental settings: intra-stimulus verification
(Experiment 1), within-DCP inter-stimulus verification
(Experiment 2), and cross-DCP inter-stimulus
verification (Experiment 3). The supplementary
results enable assessment of genuine-match performance
under a stricter false-match constraint across the
healthy and epilepsy cohorts and the four evaluated
classifiers (CRC, SVM, CNN, and BiLSTM). Together,
the two operating points provide a broader
characterization of the security--usability trade-off
than either threshold considered independently.

